\documentclass{article}
\usepackage{spconf,amsmath,graphicx,amsfonts}
\usepackage{bbm}
\usepackage[usenames,dvipsnames,svgnames,table]{xcolor}

\newcommand{\term}[1]{\textbf{#1}}
\newcommand{\cost}{\text{cost}}
\DeclareMathOperator*{\argmax}{\arg\!\max}
\DeclareMathOperator*{\argmin}{\arg\!\min}

\newcommand{\kpgcomment}[1]{\textcolor{blue}{\bf \small [ #1 --KPG]}}
\newcommand{\weiran}[1]{\textcolor{magenta}{\bf \small [ #1 --WW]}}
\newcommand{\klcomment}[1]{\textcolor{olive}{\bf \small [ #1 --KL]}}
\newcommand{\htcomment}[1]{\textcolor{red}{\bf \small [ #1 --HT]}}

\renewcommand{\kpgcomment}[1]{\ignorespaces}
\renewcommand{\weiran}[1]{\ignorespaces}
\renewcommand{\klcomment}[1]{\ignorespaces}
\renewcommand{\htcomment}[1]{\ignorespaces}

\title{End-to-end Training Approaches for Discriminative Segmental Models}

\name{Hao Tang, Weiran Wang, Kevin Gimpel, Karen Livescu}
\address{Toyota Technological Institute at Chicago\\
    \small{\texttt{\{haotang, weiranwang, kgimpel, klivescu\}@ttic.edu}}}

\begin{document}

% \ninept

\maketitle

\begin{abstract}
Recent work on discriminative segmental models has shown that they can
achieve competitive speech recognition performance, using features based
on deep neural frame classifiers.  However, segmental models can be more
challenging to train than standard frame-based approaches.  While some
segmental models have been successfully trained end to end, there is
a lack of understanding of their training under different settings and
with different losses.

We investigate a model class based on recent successful approaches,
consisting of a linear model that combines segmental features based on an
LSTM frame classifier.  Similarly to hybrid HMM-neural network models,
segmental models of this class can be trained in two stages (frame
classifier training followed by linear segmental model weight training),
end to end (joint training of both frame classifier and linear weights),
or with end-to-end fine-tuning after two-stage training.

We study segmental models
trained end to end with hinge loss, log loss, latent hinge loss, and
marginal log loss.
We consider several losses for the case
where training alignments are available as well as where they are not.

We find that in general, marginal log loss provides the most consistent
strong performance without requiring ground-truth alignments.  We also
find that training with dropout is very important in obtaining good
performance with end-to-end training.  Finally, the best results are
typically obtained by a combination of two-stage training and fine-tuning.
\end{abstract}

\begin{keywords}
Discriminative segmental models, end-to-end training
\end{keywords}

\section{Introduction}

% \begin{itemize}
% \item End-to-end approaches in speech and language
%     include CTC, encoder-decoder, lattice-free MMI training, SRNN.
% \item In vision, cite [Chen, Schwing] and [Chen, Papandreou]
% \item In theory, cite [Shalev-Shwartz, Shashua]
% \item sequence training with backprop, cite Vesely.
% \end{itemize}

End-to-end training has proved to be successful,
for example, in connectionist temporal classification (CTC) \cite{GMH2013},
encoder-decoders \cite{SVL2014},
hidden Markov model (HMM) based hybrid systems \cite{PPGGMNWK2016},
deep segmental neural networks (DSNN) \cite{ADYJ2013},
and segmental recurrent neural networks (SRNN) \cite{LKDSR2016}.
%\weiran{Shall we provide references for each of these models?}
All of these models have a feature encoder and an output
model for generating label sequences.
The feature encoder
can be a recurrent or a feedforward neural network, and the output model
can be a recurrent neural decoder, such as a long short-term memory network (LSTM),
or a probabilistic graphical model, such as an HMM,
a conditional random field (CRF),
or a semi-Markov CRF.
The actual definition of end-to-end training
is rarely made 
%never 
explicit in the literature.
In this work, we define \term{end-to-end training}
as optimizing the encoder parameters and the output model parameters
jointly. The alternative, which we refer to as \term{two-stage training}, optimizes the feature encoder and output model parameters separately in two stages. 
%, as opposed to optimizing them individually in two stages.
%Optimizing parameters individually can be seen
%as block coordinate descent, and
%the rationale behind end-to-end training is that
%\weiran{each block of parameters is optimized with respect to the most up-to-date version of the other set of parameters, so}
%loss functions can be better optimized
%when parameters are optimized jointly.
%\weiran{and end-to-end training often leads to simpler pipelines.}
%We refer to optimizing the feature encoder and the output model
%separately in two stages as \term{two-stage training}.

These two families of training approaches differ in terms of annotation requirements, computational and learning efficiency, and the loss functions customarily used for each. 
Two-stage training typically requires frame-level labels for the first stage, but may therefore require fewer samples to learn from \cite{SS2016}.  End-to-end training avoids the cascading errors of pipelines, but results in hard-to-optimize objectives that are sensitive to initialization. It is also possible to perform end-to-end fine-tuning after two-stage training, which has been found useful in past work~\cite{VGBP2013}.

In this work, we study training approaches for segmental models. 
Segmental models have been shown
to be successful when trained end to end
from scratch \cite{LKDSR2016}. We focus on
a particular class of segmental models, with LSTMs
as encoders, and linear segmental models as output models.
%\weiran{I wonder if this detail is important enough to appear in introduction? I understand that your next paragraph is related to this detail.}\kpgcomment{I made it a little more abstract}
%There is a log-softmax layer inbetween the encoder and the output model
%so that the encoded features are log probabilities.
For models trained in two stages, there is often
an extra restriction on the representation
of the encoded features. For example, they may be log 
probabilities of triphone states in HMM hybrid systems \cite{GJM2013}. 
Systems trained end to end (encoder-decoders, DSNNs, and SRNNs) are not so constrained. 
%Hidden vectors in encoder-decoder systems, DSNNs, and SRNNs are not so constrained. 
To enable fair comparison, % of training approaches, 
we use model architectures 
that seamlessly permit both kinds of training without requiring any change to 
the model parameterization. The only difference is that two-stage 
training leads to interpretable encoded features, but the functional 
architectures are identical.\footnote{We note that though our model class is suitable for studying
end-to-end systems in various aspects, 
using better encoders, such as SRNNs, might lead to better absolute performance.}
%\kpgcomment{I think we could move this paragraph to be later in the paper, and/or make it a footnote.} 
%This allows us to fairly compare training approaches %with the same model class, 
%while holding everything else fixed.

%There is evidence that performing end-to-end training
%after two-stage training can be beneficial \cite{VGBP2013}.
%However, for models trained in two stages, there is often
%an extra restriction on the representation
%of the encoded features.  For example, HMM hybrid
%systems require that the encoded features are
%log probabilities of triphone states \cite{GJM2013}, while
%hidden vectors in encoder-decoder systems, DSNNs, and SRNNs are
%not constrained to have a mapping between
%dimensions and labels.
%In general, models trained end to end from scratch
%do not have this requirement.
%Similarly in our model class, there can be a mapping
%from dimensions to labels when trained in two stages,
%but it is not required when trained end to end from scratch.
%While fixing intermediate representations may seem
%restrictive, it might require fewer samples to train \cite{SS2016},
%which further motivates us to compare two-stage training
%and end-to-end training.
\weiran{be consistent with ``end to end'' / ``end-to-end''?}

In order to thoroughly compare two-stage and end-to-end training, we consider 
a variety of loss functions and training settings. 
When end-to-end systems were first proposed, such as
CTC-LSTMs, encoder-decoders, and SRNNs,
they were tied to specific loss functions,
such as CTC, per-output cross entropy, and marginal log loss.
However, these systems can be trained with 
%many output models use the same decoding algorithm,
%and many systems can be trained with the same
%loss function.  
different loss functions; e.g., encoder-decoder systems
can be trained with hinge loss \cite{WR2016}.
%\weiran{I did not follow.}
%It is unclear whether 
It is thus important to isolate the effect
of training loss functions from models.
For our model class, the definition of encoder and
output model is completely independent of the definition
of loss functions.  This allows us to compare
training losses %against each other 
while keeping everything else fixed.

Two-stage training typically uses fine-grained labels for training the first 
stage, such as segmentations. 
%A consequence of having intermediate representations
%is that segmentations are required for training the feature encoder to produce
%such representations. 
%\weiran{I did not follow this sentence.}
For some datasets, such as TIMIT, we have the luxury
to use manually annotated segmentations, but for
most of the datasets, we do not.  If needed, 
segmentations are typically inferred by force aligning
labels to frames.  For our model class, the system
can be trained with or without segmentations
depending on the choice of loss function.

In the following sections, we explicitly define our model class and
loss functions, in particular, hinge loss and log loss
for cases where we have ground truth segmentations,
and latent hinge loss and marginal log loss
when we do not. 
%for cases where we do not have ground truth segmentations.
We perform experiments studying two-stage and end-to-end training in different settings
with different losses. 
On a phoneme recognition task, we show that end-to-end training
from scratch with marginal log loss
achieves the best result in the setting without ground
truth segmentations, while two-stage training followed by
end-to-end fine-tuning with log loss achieves the best
result in the setting with ground truth segmentations.\kpgcomment{to do: update this claim to be log loss instead of hinge once we get final results}
We also find that dropout is crucial for combating
overfitting.

\section{Discriminative segmental models}

Speech recognition, or sequence prediction in general,
can be formulated as a search problem.  The search
space is a set of paths, each of which is composed
of segments.  Each segment is associated with
a weight, and in turn each path is associated
with a weight. 
%KG changed scoring to weighted below
Prediction becomes finding
the highest weighted path in the search space.
We formalize this below.

Let $\mathcal{X}$ be the input space,
a set of sequences of frames, e.g., MFCCs
or Mel filter bank outputs.
Let $L$ be the label set, e.g., a phone set
for phoneme recognition.
A \term{segment} is a tuple $(s, t, y)$,
where $s$ is the start time, $t$ is the end time,
and $y \in L$ is the label.  Two segments $e_1, e_2$
are connected if the end time of $e_1$
is the same as the start time of $e_2$. A \term{path} is
a sequence of connected segments.
A path $p = ((s_1, t_1, y_1), \dots, (s_n, t_n, y_n))$
can also be seen as a label sequence $y = (y_1, \dots, y_n)$
and a segmentation $z = ((s_1, t_1), \dots, (s_n, t_n))$,
or simply $p = (y, z)$.
\weiran{Introduce the symbol $z$ here? I.e., $z = ((s_1, t_1), \dots, (s_n, t_n))$}

Let $E$ be the set of all possible segments.
A \term{segmental model} is a tuple $(\theta, \Lambda, \phi_\Lambda)$, where
$\theta \in \mathbb{R}^d$ is a parameter vector,
$\phi_\Lambda: \mathcal{X} \times E \to \mathbb{R}^d$ is a feature function
that uses a feature encoder parameterized by the set of parameters $\Lambda$.
We will give definitions of feature encoders and feature functions
in later sections.
With a slight abuse of notation, for a path $p = (y, z)$,
let $\phi_\Lambda(x, p) = \phi_\Lambda(x, y, z) = \sum_{e \in p} \phi_\Lambda(x, e)$.
Prediction can be formulated as
\begin{align}
\argmax_{p \in \mathcal{P}} \theta^\top \phi_\Lambda(x, p)
     = \argmax_{p \in \mathcal{P}} \sum_{e \in p} \theta^\top \phi_\Lambda(x, e),
\end{align}
where $\mathcal{P}$ is the set of all paths.
Though the output contains both a label sequence
and a segmentation, the segmentation is often disregarded
during evaluation.

Learning a segmental model amounts to finding
parameters $\theta$ and $\Lambda$ that minimize
a specified loss function.  Learning can be divided
into two cases, one with access
to ground truth segmentations, and one without.
When we have ground truth segmentations,
we receive a dataset $S = \{(x_1, y_1, z_1), \dots, (x_m, y_m, z_m)\}$
and learning aims to solve
\begin{align}
\argmin_{\theta, \Lambda} \frac{1}{m} \sum_{i=1}^m \ell(\theta, \Lambda; x_i, y_i, z_i).
\end{align}
When we do not have ground truth segmentations,
we have a dataset $S = \{(x_1, y_1), \dots, (x_m, y_m)\}$,
and learning becomes solving
\begin{align}
\argmin_{\theta, \Lambda} \frac{1}{m} \sum_{i=1}^m \ell(\theta, \Lambda; x_i, y_i).
\end{align}

\weiran{Say something about testing / decoding? For decoding, we care only about the label $y$, and not so much segmentation $z$?} \klcomment{yes, in general it could be nice to discuss somewhere on an intuitive level that it might be good to be able to disregard/marginalize out segmentations (here or in the intro or next section).}

\section{Loss Functions}

\klcomment{would be nice in this section to include general citations for the losses, and mention of their use in ASR beyond segmental models. maybe also a bit of intuition to motivate cost augmented losses.}

Since segmental models fall under
structured prediction,
any general loss function for structured prediction
is applicable to segmental models.
\weiran{Is this sentence necessary? We are specifically dealing with a sequence structure.}
In particular, we investigate hinge loss and log loss
for the case with ground truth segmentations,
and latent hinge loss and marginal log loss for
the case without ground truth segmentations.
All loss definitions $\ell(\theta, \Lambda)$ below are given in terms
of a single training sample $(x, y, z)$
where $x \in \mathcal{X}$, $(y, z) \in \mathcal{P}$.
\weiran{Shall we make $z$ depend on $y$, i.e., $z(y)$ contains the set of possible segmentation for the label $y$. Or at least mention that $z$ has to be compatible with $y$, if $y$ has five phones, z has to contain five segments to be consistent.} \klcomment{one way to do this is to make $\mathcal{Z}$ depend on $y$, and then consider $z \in \mathcal{Z}(y)$.  Then the loss is undefined if $y$ and $z$ are mismatched, but I think that's a detail we can ignore.}

\term{Hinge loss} is defined as
\begin{align}
\max_{(y', z') \in \mathcal{P}}
    \Big[ & \cost((y, z), (y', z')) - \theta^\top \phi_\Lambda(x, y, z) \notag \\
    & + \theta^\top \phi_\Lambda(x, y', z') \Big],
\end{align}
where $\cost$ is a function that measures the distance between
two paths.  \term{Log loss} is defined as
\begin{align}
-\log p(y, z | x)
\end{align}
where
\begin{align}
p(y, z | x) = \frac{1}{Z} \exp(\theta^\top \phi_\Lambda(x, y, z))
\end{align}
and $Z = \sum_{(y', z') \in \mathcal{P}} \exp(\theta^\top \phi_\Lambda(x, y', z'))$. %\kpgcomment{Added $\exp$ to definition of $Z$} 
Both hinge loss and log loss require segmentations.
\weiran{Not necessarily ``ground truth'', could be any hypothesized segmentation?}
Hinge loss has an explicit cost function \weiran{to incorporate belief on dissimilarity between paths}, while log loss does not.
In fact during prediction, hinge loss is always an upper bound of the cost function. \weiran{Is the above statement necessary?}\kpgcomment{I'm not sure if it's meaningful to think of hinge loss during prediction, since you don't have the gold output.. did you mean during training instead of during prediction?}
\htcomment{What I mean is $\cost((\hat{y}, \hat{z}), (y, z))$
is upper bounded by hinge where
$\hat{y}, \hat{z} = \argmax_{y', z'} \theta^\top \phi(x, y', z')$.
Maybe there is better way of saying it, or we can just ignore it.}
Hinge loss is non-smooth due to the $\max$ operation, while
log loss is smooth.  Both hinge loss and log loss are convex in $\theta$,
yet non-convex in $\Lambda$ if a neural network is used.
\weiran{Not crucial to mention smoothness and convexity.}

\term{Latent hinge loss} is defined as
\begin{align}
\max_{(y', z') \in \mathcal{P}}
    \Big[ & \cost((y, \tilde{z}), (y', z')) - \max_{z''} \theta^\top \phi_\Lambda(x, y, z'') \notag \\
        & + \theta^\top \phi_\Lambda(x, y', z') \Big]
\end{align}
\weiran{My suggestion is to write
\begin{align}
\max_{(y', z') \in \mathcal{P}}
    \Big[ & \cost((y, \tilde{z}), (y', z')) - \phi(x, y, \tilde{z}) \notag \\
        & + \phi(x, y', z') \Big]
\end{align}
because you give definition of $\tilde{z}$ immediately anyway.}
where $\tilde{z} = \argmax_{z'' \in \mathcal{Z}(y)} \theta^\top \phi_\Lambda(x, y, z'')$
and $\mathcal{Z}(y)$ is the set of possible segmentations of $y$.
\term{Marginal log loss} is defined as
\begin{align}
-\log p(y | x) = -\log \sum_{z \in \mathcal{Z}(y)} p(y, z | x).
\end{align}
\weiran{$z$ has to be compatible with $y$.}
Both latent hinge loss and marginal log loss do not
require ground truth segmentations.  During prediction,
latent hinge loss is also an upper bound of the cost function.
Latent hinge loss is non-smooth, while marginal log loss
is smooth.
Both latent hinge loss and marginal log loss are non-convex
in both $\theta$ and $\Lambda$.

Hinge loss training for segmental models first appeared in \cite{ZG2013},
log loss in \cite{SC2004}, and marginal log loss in \cite{ZN2009}.
For training first-pass segmental models, \cite{TWGL2015} is
the first to use hinge loss, \cite{SC2004} is the first to use log loss,
and \cite{Z2012} is the first to use marginal log loss.
For training first-pass segmental models end to end, \cite{ADYJ2013} is
the first to use marginal log loss.
Other loss functions, such as
empirical Bayes risk and structured ramp loss, have been
used in \cite{TGL2014} for training segmental models.

The above loss functions can be optimized
with stochastic gradient descent or its variants.
\weiran{change ``gradient'' to ``subgradient''?}
We propagate gradients back through the feature function
$\phi$, allowing all parameters to be updated jointly.

\section{Feature Functions}
\label{sec:feat}

Here we define explicitly feature functions we will use
in the experiments.  These feature functions
first appeared in \cite{HF2012}.

We assume there is a \term{feature encoder}, for example, an LSTM, which
produces $h_1, \dots, h_T$ given input $x_1, \dots, x_T$.
For any $t \in \{1, \dots, T\}$, we project $h_t$ to a $|L|$-dimensional vector
and pass the resulting vector through a log-softmax
layer and get $k_t$. In other words,
$k_{t, i} = \sum_j W_{ij} h_{t, j} - \log \sum_\ell \exp(\sum_j W_{\ell j} h_{t, j})$,
where $W$ is the projection matrix.
In this case, the set of parameters $\Lambda$ includes
the projection matrix $W$ and parameters in the LSTM.

\weiran{We could use words instead of equations to describe the features.}
The following is a list of features, and the final feature
function produces a concatenation of feature vectors produced
by the individual feature functions.
The average of frames over a segment is defined as
\begin{align}
\phi_{\text{avg}}(x, (s, t, y)) = \frac{1}{t - s} \sum_{i=s}^{t-1} k_i \otimes \mathbf{1}_y
\end{align}
where $\otimes$ is the tensor product,
and $\mathbf{1}_y$ is a $|L|$-dimensional one-hot vector for the label $y$.
The frame sample at the $r$-th percentile is defined as
\begin{align}
\phi_{\text{at-$r$}}(x, (s, t, y)) = k_{\lfloor s + rd \rfloor} \otimes \mathbf{1}_y
\end{align}
where $d = t - s + 1$. 
% KG changed mathbbm to mathbf above and below (since all others were mathbf, I assumed the two instances of mathbbm were mistakes)
The frame at the left boundary is defined as
\begin{align}
\phi_{\text{left-$r$}}(x, (s, t, y)) = k_{s-r} \otimes \mathbf{1}_y
\end{align}
and similarly, the frame at the right boundary is
\begin{align}
\phi_{\text{right-$r$}}(x, (s, t, y)) = k_{t+r} \otimes \mathbf{1}_y.
\end{align}

Additionally, we have features that do not depend on the feature
encoder.  The length score is defined as
\begin{align}
\phi_{\text{len}}(x, (s, t, y)) = \mathbf{1}_d \otimes \mathbf{1}_y
\end{align}
where $d = t - s + 1$.
Finally, there is a bias for each individual label
\begin{align}
\phi_{\text{bias}}(x, (s, t, y)) = \mathbf{1}_y.
\end{align}

Gradients are propagated through vectors $k_1, \dots, k_T$
to the feature encoder.  Parameters of the entire
segmental model, including the feature encoder,
can be updated jointly.

\section{Experiments}

We conduct phonetic recognition experiments on TIMIT,
a 6-hour phonetically transcribed dataset.
\weiran{Citation for TIMIT, but no need to describe ``6-hour phonetically transcribed''?}
We follow the conventional setting, training
models on the 3696-utterance training set,
and evaluate on the 192-utterance core test set.
We use the rest of the 400 utterances in the test set
as the development set.  Following the convention,
we collapse 61 phones down to 48 for training,
and further collapse them to 39 phones for
evaluation.

The feature encoder we use is a 3-layer bidirectional
LSTM with 256 cells per layer.
The outputs of the third layer are
projected from 256 dimensions to 48 and
pass through a log-softmax layer so that the final
output are log probabilities.
Inputs to the encoder are 39-dimensional MFCCs,
normalized per dimension by subtracting
the mean and dividing by the standard deviation calculated
from the training set.

\subsection{Two-Stage Training}
\label{sec:two-stage}

Since TIMIT is phonetically transcribed, we have access
to phone labels for each individual frame.  We first
train LSTM frame classifiers with cross entropy loss at
each frame.  This LSTM will serve as our feature
encoder later on, and training such LSTM corresponds
to the first stage of two-stage learning.
LSTM parameters are initialized uniformly
in the range $[-0.1, 0.1]$.  Biases for forget gates are
initialized to one \cite{JZS2015}, while other biases are initialized to zero.
Dropout for LSTMs \cite{ZSV2014} is applied at all input layers and the last
output layer with a dropout rate of 50\%.
\weiran{Mention dropout after initialization?}
We compare AdaGrad with step sizes in $\{0.01, 0.02, 0.04\}$
and RMSProp with step size 0.001 and decay 0.9.
\weiran{In the end all numbers are trained using Adagrad, right? No need to mention RMSProp.}
Mini-batch size is always one utterance. \weiran{One utterance?}
Both optimizers are run for 50 epochs.
We choose the best performing model according to the frame
error rate on the development set,
also known as early stopping.
No gradient clipping is used during training.
For comparison, following \cite{TWGL2015}
we train a convolutional neural network
(CNN) consisting of 5 layer convolution followed by
3 fully-connected layers.
Frame classification results are shown in Table~\ref{tbl:frame-exp}.
We observe that the best performing LSTM achieves a comparable frame
error rate as the CNN.  With dropout, the frame error rate
is further lowered.

\begin{table}
\caption{Frame error rates for different encoder architectures.
    \label{tbl:frame-exp}}
\vspace{-0.3cm}
\begin{center}
\begin{tabular}{llll}
         & feat     & dev      & test     \\
\hline
CNN      & MFCC+fbank
                    & 22.27    & 23.03    \\
LSTM 256x3
         & MFCC     & 22.60    & \phantom{23.20}    \\
LSTM 256x3 +dropout
         & MFCC     & 21.09    & 21.36    
\end{tabular}
\end{center}
\end{table}

After obtaining LSTM frame classifiers, we proceed to the second stage,
training segmental
models with features based on LSTM log probabilities.
Segmental models are trained with the four loss functions
for 50 epochs with early stopping.
Overlap cost \cite{TGL2014} is used in hinge loss and latent hinge loss.
A maximum duration of 30 frames is imposed.
We use feature functions described in Section~\ref{sec:feat}.
No regularizer is used except early stopping.
We compare AdaGrad with step sizes in $\{0.1, 0.2, 0.4\}$
and RMSProp with step size 0.001 and decay 0.9.
Phonetic recognition results for hinge loss are shown in Table~\ref{tbl:2-stage-hinge}. 
We observe that LSTMs perform better in frame classification,
but give little improvement over CNNs in phonetic recognition.
Recognition results for the rest of the losses are in Table~\ref{tbl:2-stage}.
Note that even though latent hinge loss and marginal log loss
do not require segmentations during training,
we do use ground truth segmentations for training
the frame classifier.  It is not a common setting,
and is done purely for comparison purposes.
We observe that, except latent hinge, other losses
perform equally well, with log loss having a slight
edge over the others.

\begin{table}
\caption{Phone error rates for segmental models trained with hinge loss
    using log probabilities generated from various encoders
    in Table~\ref{tbl:frame-exp}.
    \label{tbl:2-stage-hinge}}
\begin{center}
\begin{tabular}{llll}
         & feat     & dev      & test     \\
\hline
CNN      & MFCC+fbank
                    & 21.4     & 22.5     \\
LSTM 256x3
         & MFCC     & 23.1     & \phantom{24.2}     \\
LSTM 256x3 +dropout
         & MFCC     & 21.4     & 22.1     
\end{tabular}
\end{center}
\end{table}

% \weiran{You discussed Table 4 before Table 3 in the text.}
% To verify the usefulness of segmental models, we
% generate label sequences by removing repeated labels
% from frame-level predictions, and measure the
% phone error rates.
% This one-best decoding is similar to the one-best
% decoding in CTC-LSTMs \cite{GFGS2006}.
% Results are shown Table~\ref{tbl:frame2seq}.
% We can see that using segmental models gives
% much better performance, so it shows that
% converting frame-level log probabilities to label sequences
% is a nontrivial task.
% This concludes the two-stage training.
% \weiran{``removing repeated labels from frame-level predictions'' is a week baseline, there is no smoothing for the predicted labels. Is it necessary to mention in the paper?}

\begin{table}
\caption{Phone error rates for segmental models
    trained in two stages with different losses.
    \label{tbl:2-stage}}
\begin{center}
\begin{tabular}{lll}
      & dev      & test     \\
\hline
hinge     
%       & 21.9     & 22.5     \\
      & 21.4     & 22.1     \\
log loss 
      & 21.2     & 21.9     \\
latent hinge
      & 23.5     & 24.6     \\
marginal log loss
      & 21.6     & 22.5     
\end{tabular}
\end{center}
\end{table}

% \begin{table}
% \caption{Phone error rates by removing
%     repeated labels from frame-level prediction
%     for different encoder architectures.
%     \label{tbl:frame2seq}}
% \begin{center}
% \begin{tabular}{llll}
%          & feat     & dev    \\
% \hline
% CNN      & MFCC+fbank
%                     & 52.0   \\
% LSTM 256x3 +dropout
%          & MFCC     & 30.0
% \end{tabular}
% \end{center}
% \end{table}

\subsection{End-to-End Training with Warm Start}

After two-stage training, we fine tune the encoder and segmental model jointly
to further lower the training loss.
The four losses are compared with and without dropout.
When dropout is used, dropout rate 50\% is chosen to match the rate
during frame classifier training.
The input layers and the output layer are scaled by 0.5
when no dropout is used.
First, we initialize the models with the one trained with hinge loss above.
We run AdaGrad with step size
0.001 for 10 epochs with early stopping.
Results are shown in Table~\ref{tbl:fine-tuning-hinge}.
We observe healthy reductions
in phone error rates by fine tuning the two-stage system
across all loss functions.
We also find that fine-tuning without dropout tends to be
better than with dropout.
\weiran{But drop rate of $0.5$ might be too large, and a smaller rate may do better (as seen in Table 6).}
Though fine-tuning with hinge loss leads to the most error reduction,
we note that the two-stage system is trained with hinge loss.
At least we are centain that the two-stage system trained with hinge
loss is a descent initialization for other losses.
\weiran{Ideally, log-loss end-to-end shall be initialized with log-loss two-stage, etc. The conclusion here is a bit weak.}

Minimizing other losses from a model trained with hinge loss
is less than ideal.  We repeat the above experiments by
warm-starting from a model trained with the
loss function that we are going to minimize.
Results are shown in Table~\ref{tbl:fine-tuning}.
We observe significant gains for log loss and marginal log loss
if initialized with the matching loss function.
Similarly, the gains with dropout in these cases are smaller
than without dropout.

% \begin{table}
% \caption{Phone error rates for segmental models trained end to end initialized from
%     the best two-stage system. \label{tbl:fine-tuning}}
% \begin{center}
% \begin{tabular}{llll}
%          & dropout    & dev        & test      \\
% \hline
% hinge    & 0          & 19.3       & 20.9      \\
%          & 0.5        & 21.1       & \phantom{22.6}      \\
% \hline
% log loss & 0          & 19.7       & \phantom{20.9}      \\
%          & 0.5        & 19.3       & 21.3      \\
% \hline
% latent hinge
%          & 0          & 19.4       & 21.0      \\
%          & 0.5        & 20.9       & \phantom{22.3}      \\
% \hline
% marginal log loss
%          & 0          & 20.1       & \phantom{21.9}      \\
%          & 0.5        & 20.0       & 22.1      \\
% \end{tabular}
% \end{center}
% \end{table}

\begin{table}[t]
\caption{Phone error rates for segmental models trained end to end initialized from
    the two-stage system trained with hinge loss. \label{tbl:fine-tuning-hinge}}
\begin{center}
\begin{tabular}{llll}
         & dropout    & dev        & test      \\
\hline
hinge    & 0          & 19.4       & 20.7      \\
         & 0.5        & 20.8       &           \\
\hline
log loss & 0          & 20.2       & 21.7      \\
         & 0.5        & 21.1       &           \\
\hline
latent hinge
         & 0          & 19.3       & 21.0      \\
         & 0.5        & 20.8       &           \\
\hline
marginal log loss
         & 0          & 20.7       & 22.2      \\
         & 0.5        & 20.9       &           
\end{tabular}
\end{center}
\end{table}

\begin{table}[t]
\caption{Phone error rates for segmental models trained end to end initialized from
    two-stage systems trained with corresponding loss functions. \label{tbl:fine-tuning}}
\begin{center}
\begin{tabular}{llll}
         & dropout    & dev        & test      \\
\hline
hinge    & 0          & 19.4       & 20.7      \\
         & 0.5        & 20.8       &           \\
\hline
log loss & 0          & 18.8       & 19.7      \\
         & 0.5        & 20.3       &           \\
\hline
latent hinge
         & 0          & 20.0       & 21.2      \\
         & 0.5        & 22.1       &           \\
\hline
marginal log loss
         & 0          & 19.2       & 20.8      \\
         & 0.5        & 21.0       &           
\end{tabular}
\end{center}
\end{table}

\subsection{End-to-End Training from Scratch}

\weiran{Maybe we shall divide experiments section into subsections, with different settings for each subsection: two-stage with ground truth segments, warm-start end-to-end, end-to-end without ground-truth segments}

Next, we train the same architecture end to end
from scratch.  We make sure that all the models
are initialized identically to the two-stage systems.
The four losses are used
for training with dropout rates in $\{0, 0.1, 0.2, 0.5\}$.
Ground truth segmentations are used when training with hinge loss and log loss,
and are disregarded when training with latent hinge loss and
marginal log loss.
\weiran{You are not using ground-truth segments now, correct? Then you need to mention how segmentations are obtained for hinge/log-loss, i.e., forced-alignment from MLL.}
The optimizers we use here are
SGD with step sizes in $\{0.1, 0.5\}$, momentum 0.9, and gradient clipping at norm 5,
AdaGrad with step sizes in $\{0.01, 0.02, 0.04\}$ and
no clipping, and RMSProp with step size 0.001, decay 0.9,
and no clipping.  We run earch optimizer for 50 epochs
with early stopping.  Results are shown in Table~\ref{tbl:e2e}.
First, all optimizers above fail to minimize
latent hinge loss.  All of them get stuck in local optima,
and fail to produce reasonable forced alignments.
Even though all loss functions in end-to-end training are nonconvex,
latent hinge loss is more sensitive to initialization
than other losses.
The second observation is that adding dropout
improves performance.  However,
using the same dropout rate as the two-stage system
results in worse performance.  Finally,
though behind the best fine-tuned model,
marginal log loss with dropout 0.2 slightly
edges over other losses.

\begin{table}
\caption{Phone error rates for segmental models trained end to end with dropout.
    \label{tbl:e2e}}
\begin{center}
\begin{tabular}{llll}
         & dropout  & dev      & test     \\
\hline
hinge    & 0        & 23.1     & \phantom{25.0}     \\
         & 0.1      & 22.4     & \phantom{25.7}     \\
         & 0.2      & 22.3     & 23.7     \\
         & 0.5      & 28.9     & \phantom{30.9}     \\
\hline
log loss & 0        & 24.8     & \phantom{26.9}     \\
         & 0.1      & 22.4     & \phantom{23.8}     \\
         & 0.2      & 20.8     & 22.2     \\
         & 0.5      & 22.3     & \phantom{23.2}     \\
\hline
latent hinge
         &          & failed   &          \\
\hline
marginal log loss
         & 0        & 25.3     & \phantom{27.6}     \\
         & 0.1      & 22.1     & \phantom{24.8}     \\
         & 0.2      & 20.0     & 22.0     \\
         & 0.5      & 22.0     & \phantom{23.4}     
\end{tabular}
\end{center}
\end{table}

\section{Discussion}

We have seen that end-to-end training initialized
with a two-stage system leads to the best results.
Since in end-to-end training, the meaning
of the intermediate representations is not enforced
anymore, it is unclear how the intermediate
representations deviate from the learned ones.
To answer this, we measure per-frame cross entropy for
the LSTM frame classifier after end-to-end training.
Results are shown in Table~\ref{tbl:ce}.
First, the per-frame cross entropy for the best
performing LSTM on the training set can be
as low as 0.06, which shows that a 3-layer
bidirectional LSTM with 256 cells per layer
is able to essentially memorize the entire TIMIT dataset.
However, it is severely overfitting.
Early stopping and dropout help balance
cross entropies on the training set and 
development set.  In addition,
the cross entropies on both sets drop
after end-to-end training.  It shows
that the meaning of the intermediate representations
is still maintained by the LSTMs
after end-to-end training.

\begin{table}
\caption{Average cross entropy over frames before and after end-to-end fine-tuning.
    \label{tbl:ce}}
\begin{center}
\begin{tabular}{llll}
LSTM                 & train CE & dev CE  & dev PER \\
\hline
256x3 (best train)   & 0.0569   & 2.2395  &         \\
256x3 (best dev)     & 0.4179   & 0.9442  &         \\
256x3 +dropout       & 0.4595   & 0.7466  & 21.4    \\
256x3 +dropout +e2e  & 0.3864   & 0.6928  & 19.4    
\end{tabular}
\end{center}
\end{table}

Next, since the system trained with marginal log loss
does not use the ground truth segmentations, and since
the evaluation measure (phone error rate) does not consider
segmentations, we do not know if the system is
able to discover reasonble phone boundaries without supervision.
We approach this question by aligning the label
sequences to the acoustics, and compare the resulting segmentations
against the manually annotated segmentations.
The alignment quality for different tolerance values is shown in Table~\ref{tbl:align}.
Though the results are behind models trained specifically to align \cite{KSSC2007},
the segmental model trained with marginal log loss is not supervised with
any ground truth segmentations.
Limiting the maximum duration to 30 frames also
affects the alignment performance.

\begin{table}
\caption{Forced alignment quality on the test set as a percentage of correctly positioned phone boundaries within a predefined tolerance, measured with the best-performing segmental model trained with marginal log loss.
    \label{tbl:align}}
\begin{center}
\begin{tabular}{llll}
$t \leq 10$ms & $t \leq 20$ms & $t \leq 30$ms & $t \leq 40$ms \\
\hline
64.5          & 86.8          & 94.7          & 96.7   
\end{tabular}
\end{center}
\end{table}

Since most speech datasets do not have manually annotated
segmentations, it is desirable to train without manual alignments.  As we now know, the alignments produced by our
system trained with marginal log loss are of good quality.  Therefore, we
can use the forced alignments to train a two-stage system
followed by end-to-end fine-tuning.
We follow the exact same procedure as in the previous two-stage experiments
by training an LSTM frame classifier with the forced alignments,
followed by training a segmental model with hinge loss.
The frame error rate on the development set of the LSTM classifier
is 21.68\% against the forced alignments
and 28.91\% against the ground-truth segmentations.
Though the frame error rate is significantly worse than
when training with ground-truth segmentations,
this two-stage system achieves a phone error rate of
21.0\% on the development set.
We then fine-tune the entire system with hinge loss.
The final system achieves 18.6\% phone error rate on the development set,
and 20.1\% on the test set, a significant improvement
from the model trained end-to-end with marginal log loss, 
while not relying on ground truth segmentations.

% \begin{table}
% \caption{Frame error rates for LSTMs training on forced alignments
%     against different segmentations.}
% \begin{center}
% \begin{tabular}{lll}
%                        & dev FER \\
% \hline
% against alignments     & 21.68-  \\ 
% against ground truths  & 28.91?  \\ 
% \end{tabular}
% \end{center}
% \end{table}

% \begin{table}
% \caption{Phone error rates for segmental models trained with forced alignments.}
% \begin{center}
% \begin{tabular}{llll}
%          & dev      & test     \\
% \hline
% hinge    & 21.0-    &          \\
% log loss &          &          \\
% latent hinge
%          &          &          \\
% marginal log loss
%          &          &          
% \end{tabular}
% \end{center}
% \end{table}

In terms of efficiency in training,
all four losses require forward-backward-like algorithms
for computing gradients.  Hinge loss requires one pass on
the entire search space, log loss requires two passes on the entire search space,
latent hinge requires one pass on the entire search space and
one on the segmentation space, and marginal log loss
requires two passes on the entire search space and
two passes on the segmentation space.
The average number of hours per epoch spent on computing
gradients, excluding LSTM computations,
is shown in Table~\ref{tbl:time}.
To put them into context, feeding forward
and backpropagation for LSTMs takes 1.65 hours per epoch.
The timing is done on a single 3.4GHz four-core CPU.
The number of hours is consistent with the number of passes
required to compute gradients.
Note that the time spent on LSTMs can be halved without
incurring a performance loss by applying
frame skipping \cite{MLWZG2015, VDH2013} as shown for segmental models in \cite{TWGL2016}.

\begin{table}
\caption{Average number of hours per epoch spent
    on computing gradients excluding LSTM computations.
    \label{tbl:time}}
\begin{center}
\begin{tabular}{llll}
hinge & log loss & latent hinge & marginal log loss \\
\hline
0.52  & 1.08     & 0.73         & 2.10
\end{tabular}
\end{center}
\end{table}

\section{Conclusion}

In this work, we study end-to-end training in the context of segmental models.
The model class of choice includes a 3-layer bidirectional LSTM as feature
encoder and a segmental model using the features to produce label sequences.
This model class is suitable for studying end-to-end training, due
to its flexibility to be trained either in a two stage manner, or end to end.
The hypothesis is that training such systems in two stages is easier than
end-to-end training from scratch.  On the other hand, 
%flip side of the coin,
end-to-end training can better optimize the loss function, but it might
be sensitive to initialization.

Our model definition is separated from the definition of loss functions,
giving us the flexibility to choose loss functions based on
the training settings.
We consider two common training settings, one with ground truth
segmentations and one without.  Hinge loss and log loss
require segmentations by definition, while latent hinge loss
and marginal log loss do not.

We show that in the case where we have ground truth segmentations,
two-stage training followed by end-to-end training is significantly
better than two-stage training alone (improving upon it by 10\% relative) and end-to-end training from scratch. 
In addition, we find that end-to-end training with marginal log loss
from scratch achieves competitive results.  As a byproduct,
the system is able to generate high-quality forced alignments.
To remove the dependency on ground truth segmentations,
we train another model on the forced alignments in two stages
followed by end-to-end fine-tuning, improving upon end-to-end training from scratch by 8.6\% relative. 
The final product
is a strong system trained end to end without requiring
ground truth segmentations.

%In summary, end-to-end training with warm start can
%improve upon two-stage training by 10\% relative.
%Two-stage training followed by end-to-end training
%with forced alignments can improve upon end-to-end training
%from scratch by 8.6\% relative.

\section{Acknowledgement}

This research was supported by a Google faculty research award
and NSF grant IIS-1433485.  The opinions expressed in this
work are those of the authors and do not necessarily reflect the
views of the funding agency.  The GPUs used for this research
were donated by NVIDIA.

\bibliographystyle{IEEEbib}
\bibliography{refs}

\clearpage

\end{document}